\RequirePackage{fix-cm}
\documentclass[smallextended]{svjour3}       % onecolumn (second format)

\smartqed  % flush right qed marks, e.g. at end of proof
\usepackage{graphicx}
\usepackage{amssymb}
\usepackage{booktabs}
\usepackage{tabularx}
\usepackage{array}
\usepackage{lscape}
\usepackage{longtable}
\usepackage{multirow}
\usepackage{rotating}
\usepackage{amsmath}
\usepackage{colortbl}
\usepackage{epsfig}
\usepackage{epsf}
\usepackage{subfigure}
\usepackage{textcomp,booktabs}
\journalname{Applied Intelligence}
\begin{document}

\title{A New Medical Diagnosis Method Based on Z-Numbers}


\author{Dong Wu         \and
        Xiang Liu \and
        Feng Xue \and
        Hanqing Zheng \and
        Yehang Shou \and
        Wen Jiang%etc.
}


\institute{
            Corresponding author: Wen Jiang \at
            School of Electronics and Information, Northwestern Polytechnical University, Xi'an, Shaanxi, 710072, China\\
            Tel.: +86-29-88431267\\
              \email{jiangwen@nwpu.edu.cn}
              \and
              Dong Wu \at
              School of Electronics and Information, Northwestern Polytechnical University, Xi'an, Shaanxi, 710072, China \\
                      %  \\
              \and
           Xiang Liu \at
              Infrared Detection Technology Research \& Development Center, Shanghai Institute of Spaceflight Control Technology, CASC, Shanghai, China\\
              Shanghai Institute of Spaceflight Control Technology, Shanghai 200233, China
              %\email{smartbirdlx@gmail.com}
            \and
            Feng Xue \at
            School of Electronics and Information, Northwestern Polytechnical University, Xi'an, Shaanxi, 710072, China
              %\email{xue\_feng2006@nwpu.edu.cn}
            \and
            Hanqing Zheng \at
           Shanghai Institute of Spaceflight Control Technology, CASC, Shanghai, China
             %\email{14863278@qq.com}
            \and
            Yehang Shou \at
            School of Electronics and Information, Northwestern Polytechnical University, Xi'an, Shaanxi, 710072, China
}

\date{Received: date / Accepted: date}

\maketitle

\begin{abstract}
How to handle uncertainty in medical diagnosis is an open issue. In this paper, a new decision making methodology based on Z-numbers is presented. Firstly, the experts' opinions are represented by Z-numbers. Z-number is an ordered pair of fuzzy numbers denoted as $Z = (A, B)$. Then, a new method for ranking fuzzy numbers is proposed. And based on the proposed fuzzy number ranking method, a novel method is presented to transform the Z-numbers into Basic Probability Assignment (BPA). As a result, the information from different sources is combined by the Dempster' combination rule. The final decision making is more reasonable due to the advantage of information fusion. Finally, two experiments, risk analysis and medical diagnosis, are illustrated to show the efficiency of the proposed methodology.
\keywords{Medical diagnosis \and Z-numbers \and Fuzzy numbers \and Decision making \and Dempster-Shafer evidence theory \and Ranking Z-numbers \and Risk assessment}
\end{abstract}

\section{Introduction}
\label{intro}
With the development of society, a lot of harmful substances have affected human health, which leads to a high probability of human diseases. Therefore, medical diagnosis \cite{Wang2016A} is particularly important. However, there is still a serious lack of effective methods in addressing medical diagnostic problems. As a result, how to realize effectively medical diagnosis is still an open issue.

Medical diagnosis belongs to the application of computers in decision-making and artificial intelligence. Up to now, the study on medical diagnosis has been done by many scholars \cite{Wang2016A,Hajarolasvadi2016Employing,Das2016Medical,Juan2014Agent}. In $2016$, Kathryn Z \cite{Smith2016Past} examined the relationship between Opioid Use Disorder diagnosis, PTSD diagnosis with NMOU, and average monthly frequency of NMOU. Woolard \emph{et al.} \cite{Woolard2016A} introduced a retrospective study to show the extent of compliance with perioperative guidelines in patients. In \cite{Ahn2012A}, the recent development of mobile detection instruments used for medical diagnosis was reviewed. The features of GGT in patients that improve diagnosis efficiency were tried to unravel in \cite{Wang2016The}. However, these methods above do not take into account fuzzy concept and uncertainties of medicine. In fact, due to the own characteristics of medicine \cite{Ma2015Parameters,2015Toward,Deng2015Newborns}, more fuzzy concept and more uncertainties, some mathematics methods, which have the ability to deal with the fuzzy and uncertain information, are needed for solving medical diagnosis problems. Recently, fuzzy mathematics \cite{Zadeh1965Fuzzy} and Dempster-Shafer (DS) evidence theory \cite{Shafer1976A,Dempster1967Upper,fu2015group,Wang2016A} are widely applied in medical diagnosis, since they could reasonably model uncertainty and fuzzy information and describe them. Wang \emph{et al.} \cite{Wang2016A} adopted fuzzy soft sets based on ambiguity measure and Dempster-Shafer evidence theory, and the method was applied in medical diagnosis. Recently, a theoretical model \cite{Mak2015A} has been created to calculate the probabilities of hypothetical patients having designated diseases. And then based on the theoretical model, a fuzzy probabilistic method was presented to estimate the probability of a patient having a certain disease. In \cite{Michalski2011Generalized}, an application of GIFSS that was defined by Michalski demonstrated through a practical example of a multi-criteria medical diagnosis problem. Fuzzy soft set theory was applied through well-known Sanchez's approach for medical diagnosis using fuzzy arithmetic operations \cite{2013Fuzzy}. In \cite{Chen2013The}, an extended QUALIFLEX approach for dealing with a medical decision-making problem \cite{Semler2015Leveraging} in the context of interval type-$2$ fuzzy sets was proposed. Jose \cite{Kwon2016Medical} proposed a new methodology to combine fuzzy rule-based classification systems with interval-valued fuzzy sets \cite{Lee2012Fuzzy}, which is a suitable tool to face the medical diagnosis. An approach which combines intuitionistic trapezoidal fuzzy numbers with inclusion measure for medical diagnosis was proposed by Wang \cite{Wang2015The}. Yang \cite{Yang2015Data} used a linear regression model based on trapezoidal fuzzy numbers to predict which readings in the outlying data vector are suspected to be faulty for medical diagnosis. Some similarity measures \cite{Chou2016A,Song2015A} which can applied in medical diagnosis are proposed. In \cite{Muthukumar2015A}, a weighted similarity measure on intuitionistic fuzzy soft sets for medical diagnosis are presented.

In fuzzy mathematics \cite{Zadeh1965Fuzzy,Wang2016AP}, fuzzy numbers \cite{Giachetti1997Analysis,Yager1978,Jiang2015}, could describe human perception and subjectivity and could be able to handle uncertain or imprecise information \cite{dengadeng} to some extent. But in the actual research, we found that reliability of information in decision environment such as medical diagnosis, is very important too. If only to rely on the function of the fuzzy numbers, there may exist the limitation to appropriately describing reliability of information. In order to solve this problem, Zadeh \cite{Zadeh2011A} extended the concept of fuzzy numbers via introducing a new concept of Z-number. Z-number, a $2$-tuple fuzzy number, includes the restriction of the evaluation and the reliability of human judgement. Z-number, which contains two components, is an ordered pair of fuzzy numbers \cite{Zadeh2011A}. The first fuzzy number is used to represent the uncertain information in evaluation, and the second fuzzy number is used to measure the reliability or confidence in truth or probability. Therefore, Z-number can describe the level of human judgment and can be more effectively applied in decision-making such as medical diagnosis, fault diagnosis \cite{Yang2015Datam}.

However, in the procedure of applying Z-numbers such as decision-making, we have to face an issue, that is how to address the restriction and the reliability of Z-number \cite{Zadeh2011A}. Up to now, the study on Z-number has been done by some scholars. Ever since Kang \emph{et al.} \cite{Kang2012A} proposed an approach to convert Z-numbers into fuzzy numbers, in which the second component is defuzzified to a crisp number, numerous researchers proposed some useful methods to deal with the problems in the uncertain environment by applying the approach \cite{Kang2012A}. In order to address linguistic decision making problems, Kang \emph{et al.} \cite{Kang2012Decision} presented a MCDM method with Z-numbers based on the method introduced in \cite{Kang2012A}. Bakar \cite{Bakar2015Multi} introduced a multi-layer method to rank Z-numbers, in which there are two layers, namely, Z-number conversion and fuzzy number ranking. In \cite{Mohamad2014A}, Mohamad \emph{et al.} proposed a decision making procedure based on Z-numbers, in which Z-numbers are first transformed to fuzzy numbers, and then a ranking fuzzy number method is later used to prioritize the alternatives. In all of the above described methods, a Z-number is transformed into a fuzzy number or a generalized fuzzy number via converting the second component to a crisp number. However, according to \cite{Aliev2015The}, to convert the second component to a crisp number may lead to the loss of original information, and will exist an unreasonable situation, that is, two different Z-numbers may be converted to the same fuzzy number. Obviously, some existing methods for addressing Z-numbers still have some weaknesses, and how to deal with the relationship of restriction and the reliability of Z-number has still not been effectively processed. To address this issue, in the paper, a new ranking fuzzy numbers method is firstly introduced to process Z-numbers. Then based on the new ranking method, the BPAs of Z-numbers can be generated, in which the different importance of the first component and the second component of a Z-number is considered to make the results more reliable. Finally, in order to handle the problem of lack of information, Dempster's combination rule \cite{Dempster1967Upper,Shafer1976A}, as a powerful mathematical tool for dealing with incomplete information, is applied to fuse the obtained BPAs to make the final decision. In the proposed decision-making method, instead of converting the second component of Z-number into a crisp number, we consider the first component and second component as two independent fuzzy numbers,which can reduce the lack of information. To get more effective and reliable diagnosis results in medical diagnosis, a new medical diagnosis method based on the proposed decision-making method is proposed in this paper, where Z-numbers are used to represent the medical diagnostic information.

The remainder of this paper will be organized as follows. In Section \ref{Prelim}, some definitions and concepts are introduced. In Section \ref{Defuzz}, a new ranking method for fuzzy numbers is proposed. In Section \ref{Thepro}, a new method to determine BPA based on Z-numbers is proposed. In Section \ref{Applic}, the proposed method is applied in medical diagnosis. In Section \ref{Conclu}, the conclusions are made.
\section{Preliminaries}
\label{Prelim}
In this section, some concepts used in this paper will be introduced.
\subsection{Generalized trapezoidal fuzzy number \cite{Sanguansat2011,Chen2009}}
%\begin{definition}
In a universe of discourse $X$, the membership function $\mu(x)$ of a fuzzy number maps each element $x$ in $X$ to a real interval $[0,1]$. The membership function $\mu_A(x)$ of a generalized trapezoidal fuzzy number $A = (a,b,c,d;w)$ is defined as follows:
\begin{equation}\label{gtfn}
    \mu_{A}(x) = \left\{
        \begin{array}{ll}
            0, & x <a,  \\
            \frac{{(x - a)}}{{(b - a)}}, & a \leq x\leq b,  \\
           w, & b \leq x\leq c,  \\
           \frac{{(x - c)}}{{(d - c)}}, & c \leq x\leq d,  \\
            0,                   & x >d.
       \end{array}
    \right.
\end{equation}
%\end{definition}
\subsection{Z-number \cite{Zadeh2011A}}
\begin{figure}
  \centering
  \includegraphics[scale=0.6]{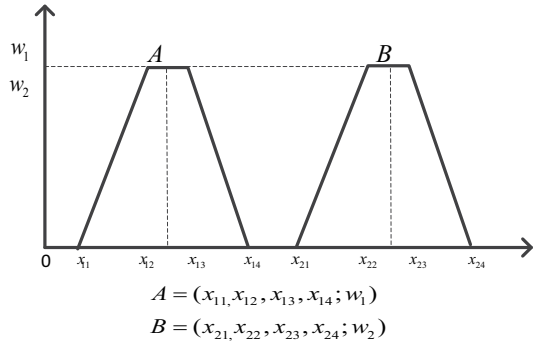}\\
  \caption{Z-number $Z = (A,B)$}\label{Zn}
\end{figure}

%\begin{definition}
A Z-number $Z = (A,B)$, shown as Fig. \ref{Zn}, contains two components. The first component $A$ is a restriction on the values which $X$ can take. The second component $B$ is a measure of reliability of the first component $A$. According to \cite{Zadeh2011A}, $(A,B)$ is an ordered pair of fuzzy numbers. Typically, $A$ and $B$ are described in a natural language \cite{Zadeh2011A}. Example: (about $37$ min, very-low).

In this paper, in order to properly deal with the problems in the uncertain environment such as medical diagnosis, the first component $A$ and the second component $B$ are described in natural language applying $9$-member linguistic terms. $9$-member linguistic terms are shown in Table \ref{9mL}.
\begin{table}
  \centering
  \caption{$9$-member Linguistic Terms for two components of Z-number}\label{9mL}
  \begin{tabular}{ccc}
    \hline
    \toprule
    Linguistic Terms & the first component & the second component \\
    \midrule
    Absolutely-low & (0,0,0,0;1.0) & (0,0,0,0;1.0) \\
    Very-low & (0,0,0.02,0.07;1.0) & (0,0,0.02,0.07;1.0) \\
    Low & (0.04,0.1,0.18,0.23;1.0) & (0.04,0.1,0.18,0.23;1.0) \\
    Fairly-low & (0.17,0.22,0.36,0.42;1.0) & (0.17,0.22,0.36,0.42;1.0) \\
    Medium & (0.32,0.41,0.58,0.65;1.0) & (0.32,0.41,0.58,0.65;1.0) \\
    Fairly-high & (0.58,0.63,0.80,0.86;1.0) & (0.58,0.63,0.80,0.86;1.0) \\
    High & (0.72,0.78,0.92,0.97;1.0) & (0.72,0.78,0.92,0.97;1.0) \\
    Very-high & (0.93,0.98,1.0,1.0;1.0) & (0.93,0.98,1.0,1.0;1.0) \\
    Absolutely-high & (1.0,1.0,1.0,1.0;1.0) & (1.0,1.0,1.0,1.0;1.0) \\
    \bottomrule
    \hline
  \end{tabular}
\end{table}

%\begin{figure}
%  \centering
%  \includegraphics[scale=0.6]{Linguistic.eps}\\
%  \caption{Linguistic Terms Representation}\label{LTR}
%\end{figure}

For example, an expert diagnosed a patient. The expert's diagnoses are represented by Z-numbers, shown as follows:

$Z_1 = (High \ probability \ that \ the \ patient \ has \ caught \ a \ Common-cold,\\ Very-high)=(High, Very-high),$

$Z_2 = (Low \ probability \ that \ the \ patient \ suffering \ from \ Meningitis,\\ Very-high)=(Low, Very-high),$

$Z_3 = (Absolutely-low \ probability \ that \ the \ patient \ suffering \ from \\\ Measles, Very-high)=(Absolutely-low, Very-high).$

According to Table \ref{9mL}, the diagnoses above can be described as follows:

$Z_1 = ((0.72,0.78,0.92,0.97;1.0),(0.93,0.98,1.0,1.0;1.0)),$

$Z_2 = ((0.04,0.1,0.18,0.23;1.0),(0.93,0.98,1.0,1.0;1.0)),$

$Z_3 = ((0,0,0,0;1.0),(0.93,0.98,1.0,1.0;1.0)).$

%\begin{table}
%  \centering
%  \caption{Linguistic Values of the result of diagnosis made by $E3$}\label{LV1}
%  \begin{tabular}{cccccccc}
%        \toprule
%                &  & & & & &  & Common-cold \\
%        \midrule
%        $E3$ & & & & & & & $Z = (A,B)$ \\
%                 & & & & & & & $= (High,Medium)$ \\
%        \bottomrule
%  \end{tabular}
%\end{table}
% In Table \ref{LV1}, $A$ denotes the degree of certainty of diagnosis made by $E3$, and $B$ denotes the measure of reliability of $A$. According to Table \ref{9mL} and Table \ref{LV1}, the diagnosis can be represented as follows:\\
%$Z = ((0.72, 0.78, 0.92, 0.97; 1.0),(0.32, 0.41, 0.58, 0.65; 1.0))$.

\subsection{Dempster-Shafer evidence theory \cite{Shafer1976A,Dempster1967Upper}}
%\begin{definition}
Let $\Theta$ be the frame of discernment: $$\Theta = \{d_1,d_2,...,d_i,...,d_n\}.$$

The BPA of $d_i$ meets the following conditions:
$$m: 2^\Theta \rightarrow [0,1],$$
$$\sum\limits_{{d_i} \subset \Theta } {m({d_i})}  = 1,$$
$$m(\emptyset)=0.$$

The BPA represents the degree of evidence support for the
proposition of $d_i$ in recognition framework. For example, $m(\emptyset)$ represents the degree of evidence support for empty set.

The Dempster's combination rule is defined as follows:
\begin{equation}\label{DS}
m(A) = \frac{1}{{1 - k}}\sum\limits_{B \cap C = A}^{} {{m_1}(B)} {m_2}(C),
\end{equation}
%\end{definition}
where $k = \sum\limits_{B \cap C = \emptyset }^{} {{m_1}(B)} {m_2}(C).$

The Dempster's combination rule could be effectively applied in decision-making such as medical diagnosis \cite{Wang2016A}.

\subsection{Maximal entropy model \cite{O'Hagan1988}}
The Maximal entropy method (MEM) was presented by O'Hagan \cite{O'Hagan1988} in $1988$. The MEM can get weights of parameters by solving a maximal entropy model. In this paper, MEM is applied to obtain the weights of parameters based on the different importance of them.

The maximal entropy model \cite{O'Hagan1988} can be defined as follows:
\begin{equation}\label{mem}
\begin{array}{ll}
   Maximize \ Disp(w) =  - \sum\limits_{i = 1}^n {{w_i}} \ln {w_i}, \\
   S.t. \ orness(w) = \alpha = \frac{1}{n - 1}\sum\limits_{i = 1}^n {(n - i)} {w_i},0\leq \alpha \leq 1, \\
   \sum\limits_{i = 1}^n {{w_i}}  = 1.
   \end{array}
\end{equation}
%\end{definition}
where ${w_i} \in \left[ {0,1} \right],i = 1,...,n.$

\section{The proposed method for ranking fuzzy numbers}
\label{Defuzz}
In many applications of Z-numbers, such as decision-making, how to deal with Z-numbers is an important issue. In this paper, in the procedure of handling Z-numbers, ranking fuzzy numbers \cite{Yager1978,Murakami1983,Chen2009,Bakar2014,jiang2016Ranking} becomes an important process. In this part, a new ranking fuzzy numbers method is proposed for dealing with Z-numbers.

\begin{figure}
  \centering
  \includegraphics[scale=0.6]{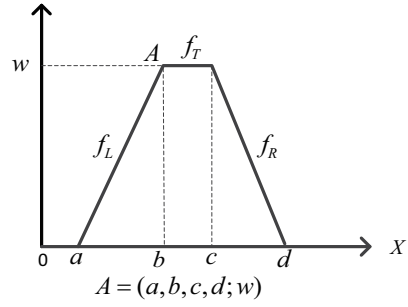}\\
  \caption{Fuzzy number $A = (a,b,c,d;w)$}\label{Fn}
\end{figure}

Firstly, the three scoring factors (or three factors affecting score) of the fuzzy number, i.e., the defuzzified value, height and spread, are calculated. Then, considering the different importance of the three scoring factors, different weights are assigned to them. Finally, ranking score of fuzzy number is defined, which reflects ranking order of the fuzzy number. Let $A$ be a fuzzy number \cite{Wang2006On,Chen2012},  $A = (a,b,c,d;w)$, as shown in Fig. \ref{Fn}, where $a$, $b$, $c$ and $d$ are real values, $w$ denotes the height of the fuzzy number $A$, and $w\in[0,1]$. The proposed method for ranking fuzzy numbers is now shown as follows:
\begin{description}
\item[Step 1:] The defuzzified value $x_A$, height $h_A$ and spread $STD_A$ of fuzzy number $A$ are calculated separately, described as follows:
    \begin{equation}\label{x_A}
{x_A} = \frac{{\int_a^b {x{f_L}dx + \int_b^c {x{f_T}dx + } \int_c^d {x{f_R}dx} } }}{{\int_a^b {{f_L}dx + \int_b^c {{f_T}dx + } \int_c^d {{f_R}dx} } }},
    \end{equation}

    \begin{equation}\label{h_A}
    h_A = w,
    \end{equation}

    \begin{equation}\label{STD_A}
ST{D_A} = \sqrt {\frac{{\sum\nolimits_{j = 1}^4 {{{(x_j - \bar x)}^2}} }}{{4 - 1}}},
    \end{equation}
where $
\bar x = \frac{{a + b + c + d}}{4}, {f_L} = \frac{{(x - a)}}{{b - a}},{f_T} = w,{f_R} = \frac{{(x - c)}}{{d - c}}.$

\item[Step 2:] Define a vector $V$ associated with the ordered arguments.

For the three scoring factors $x_A, h_A$ and $STD_A$, the ranking order of the importance of them is $x_A >h_A >STD_A$. Accordingly, $x_A, {h_A}, \frac{1}{{1 + ST{D_A}}}$, are arranged in the order of their importance from large to small, namely, the vector $V$ is defined as follows:
\begin{equation}\label{V}
V = \left( {\begin{array}{*{20}{c}}
{{x_A}}\\
{{h_A}}\\
{\frac{1}{{1 + ST{D_A}}}}
\end{array}} \right).
\end{equation}

\item[Step 3:] Calculate the weighting vector $W$ of the three ordered elements of vector $V$  by the maximal entropy model, shown as follows:
    \begin{equation}\label{mem2}
\begin{array}{*{20}{c}}
{Maximize\begin{array}{*{20}{c}}
{}&{Disp(w) =  - \sum\limits_{i = 1}^3 {{w_i}} }
\end{array}\ln {w_i},}\\
{S.t.\begin{array}{*{20}{c}}
{}&{orness(w) = \alpha  = \frac{1}{2}\sum\limits_{i = 1}^3 {(3 - i){w_i},0 \le \alpha  \le 1,} }
\end{array}}\\
{\sum\limits_{i = 1}^3 {{w_i} = 1} },
\end{array}
    \end{equation}
where $
{w_i} \in [0,1],1 \le i \le 3,W = {[{w_1},{w_2},{w_3}]^T}.$ The values of $w_1, w_2$ and $w_3$ will be respectively assigned to the three sorted elements $x_A, {h_A},\\
\frac{1}{{1 + ST{D_A}}}$ of vector $V$ according to their importance. Based on the stated above, it can be gotten that $w_1 >w_2 >w_3$, then $0.5< \alpha< 1$ that can be known in Fig. \ref{MEM} with $i = 3$. Generally, the value of $\alpha$ is defined as the middle value of the interval $(0.5,0.9]$, namely, $\alpha = 0.7$.

\begin{figure}
  \centering
  \includegraphics[scale=.4]{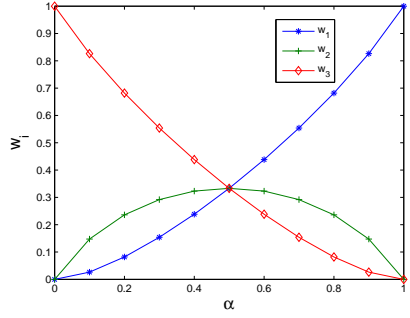}\\
  \caption{Variation of the weights with orness degree}\label{MEM}
\end{figure}

\item[Step 4:] Calculate the value of ranking score $H(A)$ of fuzzy number $A$, shown as follows:
    \begin{equation}\label{H(A)}
H(A) = {W^T}V = {w_1}{x_A} + {w_2}{h_A} + {w_3}\frac{1}{{1 + ST{D_A}}},
    \end{equation}
where $W^T$ is the transpose of weighting vector $W$. The value of score reflects the value of ranking order of the fuzzy number. The greater the value of ranking score, the better the ranking order.
\end{description}

\section{The proposed method to determine BPA}
\label{Thepro}
Suppose that a test number such as a decision $Z = (A,B) = (({x_{11}},{x_{12}},{x_{13}},{x_{14}}\\;{w_1}),({x_{21}},{x_{22}},{x_{23}},
{x_{24}};{w_2}))$. The first component of $Z$ is $A = ({x_{11}},{x_{12}},{x_{13}},\\{x_{14}};{w_1})$, and the second component of $Z$ is $B = ({x_{21}},{x_{22}},{x_{23}},{x_{24}};{w_2})$. $A$ and $B$ are two fuzzy numbers denoted in Fig. \ref{Fn}. In this section, we define the Z-number ${Z^ * }= (A^*,B^*) = ((1,1,1,1;1.0),(1,1,1,1;1.0))$ as the maximal reference number and the Z-number $Z^\Delta = (A^\Delta,B^\Delta) = ((0,0,0,0;1.0),(0,0,0,0;\\1.0))$ as the minimal reference number. Firstly, the ranking scores of two components of Z-number are calculated. Secondly, the weights of the first component $A$ and the second component $B$ are obtained by the maximal entropy model. Thirdly, the deviation degree of the test number $Z$ is defined to represent the location of the test number $Z$ between the maximal reference number $Z^*$ and the minimal reference number $Z^\Delta$. Then, the similarity measure between the test number and the reference number $Z^*$ is defined to denote the confidence degree of the test number such as a decision. Finally, the BPA is generated based on the defined deviation degree of the test number $Z$ from the reference number $Z^*$, which can be applied in decision-making such as diagnosis.

\subsection{The steps of the proposed method}
\label{Tstep}
%The method to generate the BPA of the test number $Z$ is shown as follows:
The steps of the proposed method is shown as follows:
\begin{description}
\item[Step 1:] Calculate the ranking score $H_Z(A)$ and $H_Z(B)$ of fuzzy number $A$, $B$ respectively according to Section \ref{Defuzz}.

\item[Step 2:] According to Section \ref{Defuzz}, calculate the ranking score $H_{Z^*}(A^*)$ and $H_{Z^*}(B^*)$ of fuzzy number $A^*$, $B^*$ respectively and calculate the ranking score $H_{Z^\Delta}(A^\Delta)$ and $H_{Z^\Delta}(B_\Delta)$ of fuzzy number $A^\Delta$, $B^\Delta$ respectively. The conclusions can be made as follows:
    $$H_{Z^*}(A^*) = 1, H_{Z^*}(B^*) = 1,$$
    $$H_{Z^\Delta}(A^\Delta) = 0.446, H_{Z^\Delta}(B^\Delta) = 0.446.$$
Obviously, when the value of ranking score is $1$, the ranking order is the best.

\item[Step 3:] Calculate the weights of the first component $A$ and the second component $B$ based on the maximal entropy model, shown as follows:
\begin{equation}\label{mem3}
\begin{array}{*{20}{c}}
{Maximize\begin{array}{*{20}{c}}
{}&{Disp(w) =  - \sum\limits_{i = 1}^2 {{w_i}} }
\end{array}\ln {w_i},}\\
{S.t.\begin{array}{*{20}{c}}
{}&{orness(w) = \alpha  = \sum\limits_{i = 1}^3 {(2 - i){w_i},0 \le \alpha  \le 1,} }
\end{array}}\\
{\sum\limits_{i = 1}^2 {{w_i} = 1} },
\end{array}
    \end{equation}
where $
{w_i} \in [0,1],1 \le i \le 3,\alpha  = 0.7,W = {[{w_1},{w_2}]^T}$.
The different importance of the first component $A$ and the second component $B$ is considered in this paper. According to the definition of Z-number \cite{Zadeh2011A}, the first component of Z-number is to describe the uncertainty, while the second component, as a measure of reliability of the first component, can influence but cannot decide the Z-number that can be decided by the first component. Obviously, the first component $A$ is more important than the second component $B$. Thus the first component $A$ is assigned the larger weight.

\item[Step 4:] Defined the deviation degree of the test number $Z$, shown as follows:
\begin{equation}\label{D}
D(Z) = \sqrt {\frac{{{w_1}{{({H_Z}(A) - {H_{Z^* }}(A^*))}^2} + {w_2}{{({H_Z}(B) - {H_{Z^* }}(B^*))}^2}}}{{{w_1}{{({H_{{Z^\Delta }}}(A^\Delta) - {H_{Z^* }}(A^*))}^2} + {w_2}{{({H_{{Z^\Delta }}}(B^\Delta) - {H_{Z^* }}(B^*))}^2}}}},
\end{equation}
where the deviation degree $D$ of $Z$ represents the location of the test number $Z$ between the maximal reference number $Z^*$ and the minimal reference number $Z^\Delta$, which can be shown in Fig. \ref{Locg}. Obviously, the confidence degree of the maximal reference number $Z^*$ is the largest, that is, $100\%$. The larger the value of $D$, the more far away from the reference number $Z^*$ the test number $Z$. Namely, the confidence degree of the test number Z is lower. The location of the test number Z is shown in Fig. 5. Namely, the confidence degree of the test number $Z$ is lower.

From Fig. \ref{Locg}, several conclusions can be obtained: firstly, ${Z'}^*$ represents $Z^*$ that has been processed; secondly, the location on the circle can only be on the $\frac{1}{4}$ circle at the lower left corner, since the ranking score $H_Z(A)<1$ and $H_Z(B)<1$; thirdly, when the weights of the first component $A$ and the second component $B$ are not taken into consideration, the deviation degrees of two different locations on the same circle will be same, which is unreasonable; fourthly, when the different weights are assigned to $A$ and $B$, the deviation degrees of two different locations on the same circle will be different, which is reasonable. It can be seen that the defined deviation degree is effective and reasonable.

\begin{figure}[ht!]
  \centering
  \includegraphics[scale=0.6]{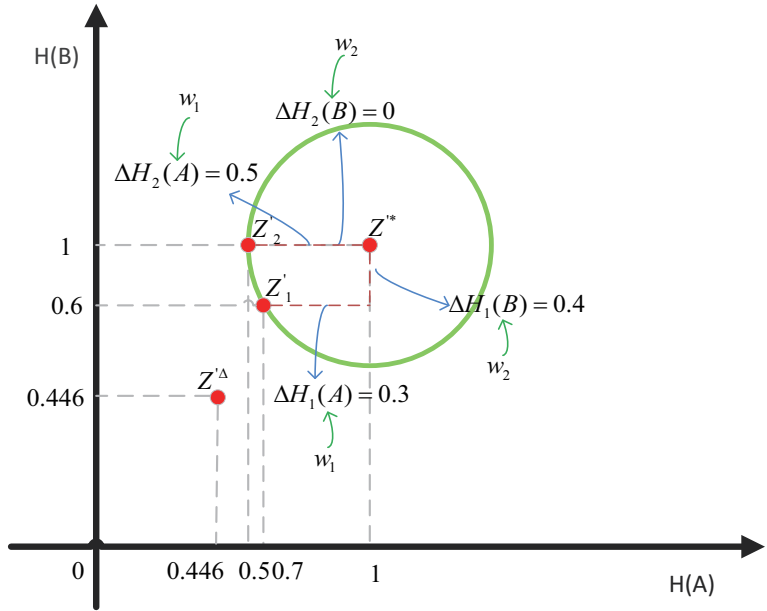}\\
  \caption{The deviation degree of the test number $Z$}\label{Locg}
\end{figure}
%The deviation degree $D$ of $Z$ is proportional to the distance between $Z$ and $Z^*$.

\item[Step 5:] Generate the BPA of the test number $Z$. Firstly, the similarity measure $S$ between the test number $Z$ such as a decision and the reference number $Z^*$ is defined, shown as follows:
    \begin{equation}\label{S}
    \begin{array}{ll}
    S &= 1-D \\
      &= {1 - \sqrt {\frac{{{w_1}{{({H_Z}(A) - {H_{Z^* }}(A^*))}^2} + {w_2}{{({H_Z}(B) - {H_{Z^* }}(B^*))}^2}}}{{{w_1}{{({H_{{Z^\Delta }}}(A^\Delta) - {H_{Z^* }}(A^*))}^2} + {w_2}{{({H_{{Z^\Delta }}}(B^\Delta) - {H_{Z^* }}(B^*))}^2}}}}}.
    \end{array}
    \end{equation}
  Obviously, the similarity measure between the test number and the reference number $Z^*$ denotes the confidence degree of the test number such as a decision. The larger the value of similarity measure, the higher confidence degree of the test number. On the other hand, the similarity measure between a test number $Z$ and the reference number $Z^*$ can also be regarded as the ranking score of $Z$ to get the ranking order of $Z$.

Then, the BPAs can be gotten by normalizing the obtained similarity measure $S$. Finally, in order to address the problem of lack of information, the obtained BPAs will be fused by Dempster's combination rule to get the final decision. In the following subsection, the proposed method will be made a comparison with the existing methods for ranking Z-numbers to illustrate the effectiveness and superiority of the proposed methodology.
\end{description}

\subsection{A comparison with the existing methods for ranking Z-numbers}
\label{Acomw}
%The relationship between the test number $Z$ and the the maximal reference number $Z^*$ is shown in Fig. \ref{Locg}. From Fig. \ref{Locg}, several conclusions can be obtained: first, ${Z'}^*$ represents $Z^*$ that has been processed; second, the location on the circle can only be on the $\frac{1}{4}$ circle at the lower left corner, since the ranking score $H_Z(A)<1$ and $H_Z(B)<1$; third, when the weights of the first component $A$ and the second component $B$ are not taken into consider, the deviation degrees or similarity measures of two different locations on the same circle will be same, which is unreasonable; four, when the different weights are assigned to $A$ and $B$, the deviation degrees or similarity measures of two different locations on the same circle will be different, which is reasonable. It can be seen that the defined deviation degree or similarity measures can be reasonably applied to handle Z-numbers.

%\begin{figure}
%  \centering
%  % Requires \usepackage{graphicx}
%  \includegraphics[scale=0.6]{Location.eps}\\
%  \caption{The location of the test number $Z$}\label{Locg}
%\end{figure}

\begin{figure}[ht!]
  \centering
  \includegraphics[scale=0.65]{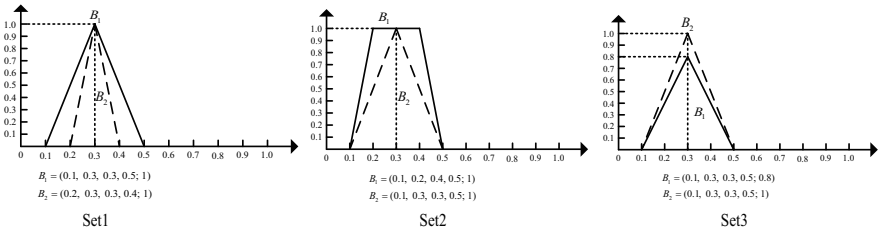}\\
  \caption{The second components of Z-numbers}\label{AllSec}
\end{figure}

In this part, the similarity measure defined in Step $5$ in subsection \ref{Tstep} is used as the ranking score of Z-number to rank Z-numbers. In the following, we use $3$ sets of Z-numbers in \cite{jiang2016Ranking} to compare the ranking results obtained by the defined similarity measure and a number of existing ranking methods to show the effectiveness and superiority of the proposed method. Let all first components $A_i$ of Z-numbers in the $3$ sets are the same, $A = (0.1,0.3,0.3,0.5;1.0)$. All the $B_i$ of Z-numbers are shown in Fig. \ref{AllSec} and the comparison results are shown in Table \ref{Comparison}. According to Fig. \ref{AllSec} and Table \ref{Comparison}, we can see the drawbacks of the existing ranking methods and the advantages of the proposed method:

  (1) If second components of two Z-numbers are $B_1$ and $B_2$ in Set $1$ of Fig. \ref{AllSec}, the ranking result, $Z_1 < Z_2$, obtained by the defined similarity measure in this paper is reasonable and consistent with human intuition, since the truth that the ranking order of two the second components is $B_1 < B_2$ according to section \ref{Defuzz} and the two first components of $Z_1$, $Z_2$ are the same. However, Mohamad's method, Bakar's method and Kang's method can't correctly address this situation and get an incorrect ranking order $Z_1 = Z_2$, since the fact that the different Z-numbers get the same ranking order. In this case, it means that the second component $B$ doesn't work in Mohamad's method, Bakar's method and Kang's method, which is not consistent with the concept of Z-number in \cite{Zadeh2011A}.

  (2) If two second components of Z-numbers are $B_1$ and $B_2$ in Set $2$ of Fig. \ref{AllSec}, the proposed method can get the correctly ranking result, $Z_1 < Z_2$, since the fact that the ranking order of two the second components is $B_1 < B_2$ according to section \ref{Defuzz} and two the first components of $Z_1$, $Z_2$ are the same. However, Mohamad's method, Bakar's method and Kang's method can't correctly address this situation and get an incorrect ranking order $Z_1 = Z_2$, since the truth that the different Z-numbers get the same ranking order. In this case, it means that in Mohamad's method, Bakar's method and Kang's method, the second component $B$ doesn't work, which is not consistent with the concept of Z-number.

  (3) If two second components of Z-numbers are $B_1$ and $B_2$ in Set $3$ of Fig. \ref{AllSec}, the ranking result, $Z_1 > Z_2$, obtained by the defined similarity measure in this paper is reasonable and consistent with human intuition, since the truth that the ranking order of the two second components is $B_1 > B_2$ according to section \ref{Defuzz} and the two first components of $Z_1$, $Z_2$ are the same. However, Mohamad's method, Bakar's method and Kang's method can't correctly address this situation and get an incorrect ranking order $Z_1 = Z_2$, since the fact that the different Z-numbers get the same ranking order. In this case, it means that the second component $B$ doesn't work in Mohamad's method, Bakar's method and Kang's method, which is not consistent with the concept of Z-number.

  In summary, from Fig. \ref{AllSec} and Table \ref{Comparison}, it is obvious that the defined similarity measure in this paper provides a reasonable ranking order and overcomes the drawbacks of the existing ranking methods. Specially, the proposed method can be applied in decision-making such as risk analysis and medical diagnosis, since it uses Z-number as a whole to model, and generates BPAs by taking into account the different importance of two components of Z-number, and can get reasonable ranking order. The above summary illustrates the effectiveness and superiority of the proposed methodology.

\begin{table}[ht!]
\renewcommand\arraystretch{0.6}
\centering\caption{A comparison with the existing methods}\label{Comparison} %\scriptsize
 \begin{tabular}{ l l l p{0.045\textwidth}l p{0.055\textwidth}l p{0.055\textwidth}l lp{0.055\textwidth}llp{0.055\textwidth}llp{0.055\textwidth}ll}
  \toprule
  \multicolumn{4}{ l }{methods}&\multicolumn{2}{ c }{Set1}&&\multicolumn{2}{ c }{Set2}&& \multicolumn{2}{ c }{Set3}&\\
  \cline{5-13}%\cline{8-9}\cline{11-12}\cline{14-15}
  \multicolumn{4}{ l }{}&$Z_1$&$Z_2$&&$Z_1$&$Z_2$&&$Z_1$&$Z_2$\\
  \midrule
  \multicolumn{4}{ l }{Mohamad's method \cite{Mohamad2014A}}&0.0774
  &	0.0774&&	0.0774&	0.0774&&	0.0774&	0.0774\\
  \multicolumn{4}{ l }{Bakar's method \cite{Bakar2015Multi}}&0.0288&	0.0288&&	0.0288&	0.0288&&	0.0288&	0.0288	\\
  \multicolumn{4}{ l }{Kang's method \cite{Kang2012Decision}}&0.3000&	0.3000&&	0.3000&	0.3000&&	0.3000&	0.3000\\
  \multicolumn{4}{ l }{The proposed method}&0.2610&	0.2663&&	0.2598&	0.2610&&	0.2278&	0.2610\\
  \bottomrule
  \end{tabular}
 \end{table}

\subsection{An illustrative experiment of application}
\label{Anex}

\begin{table}[ht!]\centering
\tiny
\caption{The evaluation of sub-components \cite{Mohamad2014A}}\label{Evalu}
    \begin{tabular}{lll}
        \toprule
        Manufactory & Sub-components & Linguistic values of
severity of loss $\tilde W_i$ \\
        \hline\\
        \multirow{3}{*}{$M_1$}   & $C_{11}$ & $\widetilde{W}_{11}=(0.12,0.24,0.24,0.36;1.0) $\\
        & $C_{12}$ & $\widetilde{W}_{12}=(0.48,0.60,0.60,0.72;1.0) $\\
        & $C_{13}$ & $\widetilde{W}_{13}=(0.0,0.12,0.12,0.24;1.0) $\\
        \hline\\
        \multirow{3}{*}{$M_2$}   & $C_{21}$ & $\widetilde{W}_{21}=(0.72,0.84,0.84,0.96;1.0) $\\
        & $C_{22}$ & $\widetilde{W}_{22}=(0.26,0.36,0.36,0.48;1.0) $\\
        & $C_{23}$ & $\widetilde{W}_{23}=(0.36,0.48,0.48,0.60;1.0) $\\
        \hline\\
        \multirow{3}{*}{$M_3$}   & $C_{31}$ & $\widetilde{W}_{31}=(0.84,1.0,1.0,1.0;1.0) $\\
        & $C_{32}$ & $\widetilde{W}_{32}=(0.0,0.0,0.0,0.12;1.0) $\\
        & $C_{33}$ & $\widetilde{W}_{33}=(0.60,0.72,0.72,0.84;1.0) $\\
        \toprule
        Manufactory & Sub-components & Linguistic values of the reliability $\tilde R_i$ \\
        \hline\\
        \multirow{3}{*}{$M_1$}   & $C_{11}$ & $\widetilde{R}_{11}=(0.24,0.36,0.36,0.48;1.0)$\\
        & $C_{12}$ & $\widetilde{R}_{12}=(0.36,0.48,0.48,0.60;1.0)$\\
        & $C_{13}$ & $\widetilde{R}_{13}=(0.48,0.60,0.60,0.72;1.0)$\\
        \hline\\
        \multirow{3}{*}{$M_2$}   & $C_{21}$ & $\widetilde{R}_{21}=(0.72,0.84,0.84,0.96;1.0)$\\
        & $C_{22}$ & $\widetilde{R}_{22}=(0.48,0.60,0.60,0.72;1.0)$\\
        & $C_{23}$ & $\widetilde{R}_{23}=(0.36,0.48,0.48,0.60;1.0)$\\
        \hline\\
        \multirow{3}{*}{$M_3$}   & $C_{31}$ & $\widetilde{R}_{31}=(0.24,0.36,0.36,0.48;1.0)$\\
        & $C_{32}$ & $\widetilde{R}_{32}=(0.6,0.72,0.72,0.84;1.0)$\\
        & $C_{33}$ & $\widetilde{R}_{33}=(0.0,0.12,0.12,0.24;1.0)$\\
        \bottomrule
    \end{tabular}
\end{table}

In order to further illustrate the effectiveness of the method, in this part, a frequently used experiment of application in decision making \cite{Mohamad2014A,Kahraman2008Fuzzy} will be done to compared the proposed method with Mohamad \emph{et al.}' method \cite{Mohamad2014A} to validate the effectiveness of the proposed method. Decision making plays an important role in our real life. Specially, decision making is the main task in the medical diagnosis. There are three manufactories $M_1$, $M_2$, and $M_3$ and each manufactory produces the components $C_1$, $C_2$, and $C_3$, respectively. A component consists of three sub-components, that is $C_{i1}, C_{i2}$ and $C_{i3}$, where $i=1,2,3$. To assess the risk faced by each sub-component, the evaluating items are represented by $\tilde W_{ik}$ and $\tilde R_{ik}$, where $1\leq k \leq 3, 1\leq i \leq 3$. $\tilde W_{ik}$ represents the severity of loss of the sub-components. $\tilde R_{ik}$ denotes the reliability of the decision maker's opinion on each sub-component. Therefore, the entries of decision matrix can be represented as $Z_{ik}(\tilde W_{ik}, \tilde R_{ik})$.

The severity of the loss of the sub-components and the reliability of the decision makers' opinion are shown in Table \ref{Evalu}.

Firstly, the risk assessment \cite{Dong2017Risk} for manufactory $C_i$ is represented by Z-numbers, which is shown in Table \ref{Rass}.

\begin{table}[ht!]\centering
\tiny
\caption{The risk assessment represented by Z-numbers}\label{Rass}
    \begin{tabular}{cc}
        \toprule
         & $M_1$ \\
        \hline\\
        \multirow{1}{*}{$C_1$}   & ${Z_{11}} = ((0.12,0.24,0.24,0.36;1.0),(0.24,0.36,0.36,0.48;1.0))$
        \\
        \hline\\
        \multirow{1}{*}{$C_2$}   & ${Z_{21}} = ((0.48,0.60,0.60,0.72;1.0),(0.36, 0.48, 0.48, 0.60; 1.0))$ \\
        \hline\\
        \multirow{1}{*}{$C_3$}   & ${Z_{31}} =
        ((0.0, 0.12, 0.12, 0.24; 1.0),(0.48, 0.60, 0.60, 0.72; 1.0))$\\

        \toprule
         & $M_2$ \\
        \hline\\
        \multirow{1}{*}{$C_1$}   & ${Z_{12}} = ((0.72, 0.84, 0.84, 0.96; 1.0),(0.72, 0.84, 0.84, 0.96; 1.0))$\\

        \hline\\
        \multirow{1}{*}{$C_2$}   & ${Z_{22}} = ((0.26,0.36,0.36,0.48;1.0),(0.48,0.60,0.60,0.72;1.0))$\\

        \hline\\
        \multirow{1}{*}{$C_3$}   & ${Z_{32}} = ((0.36, 0.48, 0.48, 0.60; 1.0),(0.36, 0.48, 0.48, 0.60; 1.0))$\\

        \toprule
         & $M_3$ \\
        \hline\\
        \multirow{1}{*}{$C_1$}   & ${Z_{13}} = ((0.84, 1.0, 1.0, 1.0; 1.0),(0.24, 0.36, 0.36, 0.48; 1.0))$
        \\
        \hline\\
        \multirow{1}{*}{$C_2$}   & ${Z_{23}} = ((0.0, 0.0, 0.0, 0.12; 1.0),(0.6, 0.72, 0.72, 0.84; 1.0))$
        \\
        \hline\\
        \multirow{1}{*}{$C_3$}   & ${Z_{33}} = ((0.60,0.72,0.72,0.84;1.0),(0.0,0.12,0.12,0.24;1.0))$
        \\
        \bottomrule
    \end{tabular}
\end{table}

Secondly, according to Section \ref{Tstep}, the BPAs of risk assessment from three sub-components are calculated based on Eqs. (\ref{mem3}-\ref{S}). The results are presented in Table \ref{TBR}.

\begin{table}[ht!]
\centering
 \caption{The BPAs of risk evaluation from three sub-components} \label{TBR}
 \tiny
 \begin{tabular}{ccccc}
   \hline
   \toprule
    & $M_1$ & $M_2$ & $M_3$ & $(M_1,M_2,M_3)$ \\
    \midrule
   $C_1$ & 0.1326 & 0.4336 & 0.3355 & 0.0983 \\
   $C_2$ & 0.3413 & 0.2571 & 0.1062 & 0.2954 \\
   $C_3$ & 0.1260 & 0.2758 & 0.2682 & 0.3300 \\
   \bottomrule
   \hline
 \end{tabular}
 \end{table}

%\begin{figure}[ht!]
%  \centering
%  % Requires \usepackage{graphicx}
%  \includegraphics[scale=0.6] {VBPA.eps}\\
%  \caption{The visualization of BPAs of risk evaluation}\label{TvBPA}
%\end{figure}
%, and the visualization of the fusion results can be shown Fig. \ref{Tfus}
Finally, the obtained BPAs are fused by Dempster's combination rule to get the final decision to address the problem of lack of information. The results denote that the manufactories $M_2$ has the highest risk or the highest probability of failure followed by $M_3$ and $M_1$.
\begin{table}[ht!]
\centering
 \caption{A comparison of the proposed method with Mohamad \emph{et al.}' method \cite{Mohamad2014A}} \label{Tfr}
 \begin{tabular}{ccccc}
   \hline
   \toprule
    methods& $M_1$ & $M_2$ & $M_3$ & $(M_1,M_2,M_3)$ \\
    \midrule
    & & & &\\
    Mohamad \emph{et al.}' method \cite{Mohamad2014A} & 0.1049 & 0.2460 & 0.0630 & 0.0000\\
    The proposed method& 0.1740 & 0.5103 & 0.2866 & 0.0291 \\
   \bottomrule
   \hline
 \end{tabular}
\end{table}

%\begin{figure}[ht!]
%  \centering
%  % Requires \usepackage{graphicx}
%  \includegraphics[scale=1]{Fusing.eps}\\
%  \caption{The visualization of fusing results}\label{Tfus}
%\end{figure}

\begin{figure}[ht!]
  \centering
  \includegraphics[scale=1]{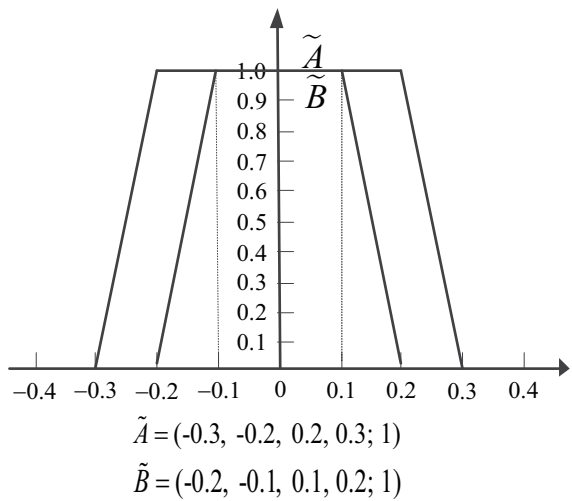}\\
  \caption{The symmetrical fuzzy numbers with $y$ axis}\label{Symme}
\end{figure}

From Table (\ref{Evalu}-\ref{Tfr}), it can be seen that compared with Mohamad \emph{et al.}' method \cite{Mohamad2014A}, the proposed method can get correct and reasonable result, that is, the ranking order of risk is $M_2>M_3>M_1$, since the truth that both sub-components $C_1$ and $C_3$ made the same assessment, that is, the manufactory $M_2$ has the highest risk and $M_1$ has the lowest risk, which illustrates the effectiveness of the proposed method. However, Mohamad \emph{et al.}' method \cite{Mohamad2014A} gets the incorrect result, $M_2>M_1>M_3$, that is, the manufactory $M_3$ has the lowest risk. On the one hand, their incorrect ranking order results from the fact that Mohamad \emph{et al.}' method does not consider the different importance of two components of Z-number, and their method converts the second component of Z-number to a crisp number, which may lead to the loss of information. On the other hand, for the three fuzzy numbers converted from the aggregated evaluation for manufactory by their method \cite{Mohamad2014A}, $M_1=(0.20, 0.32, 0.44; 0.36), M_2= (0.45, 0.56, 0.68; 0.48), M_3=(0.48, 0.57, 0.65; 0.12)$, the reasonable ranking order should be $M_2>M_3>M_1$, which is consistent with the truth that the centroid point of $M_3$ on the
$X$-axis is larger than that of $M_1$ on the $X$-axis. However, Mohamad \emph{et al.}' method obtains the incorrect ranking order $M_2>M_1>M_3$. In addition, Mohamad \emph{et al.}' method can not solve the situations that when Z-numbers are converted to the symmetrical fuzzy numbers with $y$ axis, such as the fuzzy numbers shown in Fig. \ref{Symme}. About all, it is can be seen that the proposed method can overcome the weaknesses of the previous method.

%In Fig. \ref{Locg}, the deviation degree $D_2$ of test number $Z_2$ is larger than the deviation degree $D_1$ of test number $Z_1$, namely, the distance between $Z_2$ and $Z^*$ is larger than the distance between $Z_1$ and $Z^*$.

\section{Application of the proposed method to medical diagnosis}
\label{Applic}
In this section, an application of the proposed method to medical diagnosis is illustrated. From a patient's symptoms, he may be suffering from three diseases, namely, \emph{Common-cold, Meningitis} and \emph{Measles}. There are three experts $E1$, $E2$ and $E3$, they respectively made three kinds of diagnoses \emph{(Common-cold, Meningitis, Measles)} represented by Z-numbers for the patient, which are shown in Table \ref{LV}. For example, the expert $E1$ diagnoses that the degree of certainty of \emph{Common-cold} is \emph{Very-high}, and the measure of reliability of his diagnosis is \emph{Very-high}, which can be represented by ${Z_{11}} = ({A_{11}},{B_{11}})= (Very-high, Very-high)$. The corresponding linguistic terms are shown in Table \ref{Dot}.

In Table \ref{LV}, $A$ denotes the degree of certainty of diagnosis made by the decision-makers and $B$ denotes the measure of reliability of $A$.

\begin{table}[ht!]
  \centering
  \caption{Linguistic Values of diagnoses made by three experts}\label{LV}
    \tiny
    \begin{tabular}{cccc}
        \toprule
                &   Common-Cold     &  Meningitis   &   Measles     \\
        \midrule
         $E1$       &   ${Z_{11}} = ({A_{11}},{B_{11}})$ & ${Z_{12}} = ({A_{12}},{B_{12}})$ & ${Z_{13}} = ({A_{13}},{B_{13}})$ \\
         &$=(Very-high,$ &$=(Low,$ &$=(Absolutely-low,$\\
         &$ Very-high)$     & $ Very-high)$ & $ Very-high)$ \\

        \midrule
           $E2$     &   ${Z_{21}} = ({A_{21}},{B_{21}})$ & ${Z_{22}} = ({A_{22}},{B_{22}})$ & ${Z_{23}} = ({A_{23}},{B_{23}})$ \\
         &$=(Fairly-high,$ &$=(Low,$ &$=(Low,$\\
         &$ High)$     & $ High)$ & $ Very-high)$     \\
        \midrule
            $E3$    &   ${Z_{31}} = ({A_{31}},{B_{31}})$ & ${Z_{32}} = ({A_{32}},{B_{32}})$ & ${Z_{33}} = ({A_{33}},{B_{33}})$ \\
         &$=(Low,$ &$=(Low,$ &$=(High,$\\
         &$ Very-high)$     & $ High)$ & $ Very-high)$     \\
        \bottomrule
    \end{tabular}
\end{table}

\begin{table}[ht!]
 \centering
 \caption{Diagnoses of three experts are represented by linguistic terms} \label{Dot}
    \tiny
 \begin{tabular}{ll}
   \hline
   \toprule
   expert & Z-numbers represented by linguistic terms \\
   \midrule
          &${Z_{11}} = ((0.93,0.98,1.0,1.0;1.0),(0.93,0.98,1.0,1.0;1.0))$ \\
   $E1$ &${Z_{12}} = ((0.04,0.1,0.18,0.23;1.0),(0.93,0.98,1.0,1.0;1.0))$ \\
    &${Z_{13}} = ((0,0,0,0;1.0),(0.93,0.98,1.0,1.0;1.0))$ \\
    \midrule
    & ${Z_{21}} = ((0.58,0.63,0.80,0.86;1.0),(0.72,0.78,0.92,0.97;1.0))$ \\
   $E2$ &${Z_{22}} = ((0.04,0.1,0.18,0.23;1.0),(0.72,0.78,0.92,0.97;1.0))$\\
    &${Z_{23}} = ((0.04,0.1,0.18,0.23;1.0),(0.93,0.98,1.0,1.0;1.0))$ \\
    \midrule
    &${Z_{31}} = ((0.04,0.1,0.18,0.23;1.0),(0.93,0.98,1.0,1.0;1.0))$ \\
   $E3$ &${Z_{32}} = ((0.04,0.1,0.18,0.23;1.0),(0.72,0.78,0.92,0.97;1.0))$ \\
    &${Z_{33}} = ((0.72,0.78,0.92,0.97;1.0),(0.93,0.98,1.0,1.0;1.0))$ \\
    \bottomrule
   \hline
 \end{tabular}
 \end{table}

In order to solve the problem of the loss of information and enhance the reliability of the results, the BPAs obtained in Section \ref{Thepro} are fused by Dempster's combination rule. The procedure of the proposed medical diagnosis method, as shown in Fig. \ref{lct}, is detailed as follows:

\begin{figure}[ht!]
  \centering
  \includegraphics[scale=0.4]{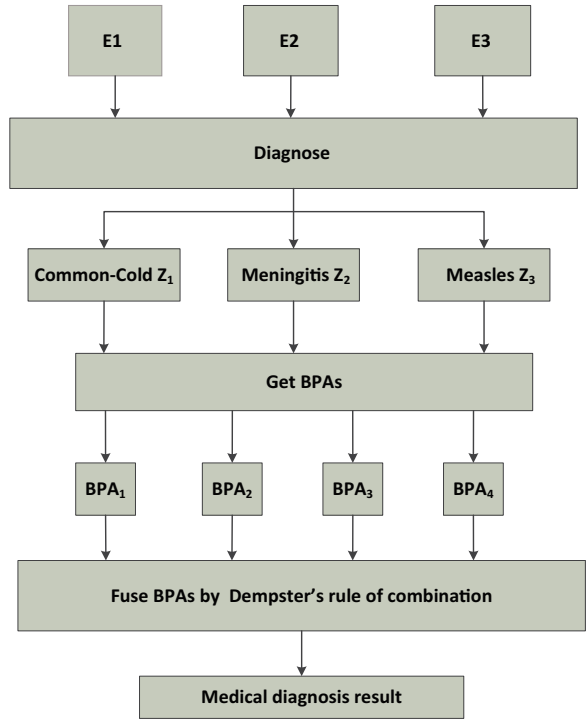}\\
  \caption{Structure of medical diagnosis}\label{lct}
\end{figure}

\begin{description}
\item[Step 1:] According to Section \ref{Defuzz}, calculate respectively the ranking score of the two components $A$ and $B$ of diagnoses $Z_1, Z_2, Z_3$ shown in Table \ref{Dot}. Take the diagnoses of $E1$ in Table \ref{Dot} as an example, shown as follows:

\begin{equation*}\label{diagnosis1}
\begin{array}{ll}
  Z(Common-Cold) &=  (A,B)\\
      &= ((0.93,0.98,1.0,1.0;1.0),(0.93,0.98,1.0,1.0;1.0)),
\end{array}
\end{equation*}

\begin{equation*}\label{diagnosis2}
\begin{array}{ll}
 Z(Meningitis) &=  (A,B)\\
      &= ((0.04,0.1,0.18,0.23;1.0),(0.93,0.98,1.0,1.0;1.0)),
\end{array}
\end{equation*}

\begin{equation*}\label{diagnosis3}
\begin{array}{ll}
  Z(Measles) &=  (A,B)\\
      &= ((0,0,0,0;1.0),(0.93,0.98,1.0,1.0;1.0)).
\end{array}
\end{equation*}

\begin{figure}
  \centering
  \includegraphics[scale=0.6]{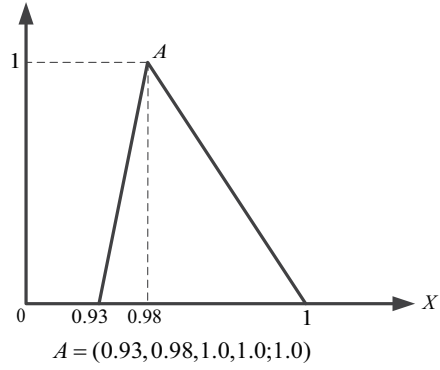}\\
  \caption{The first component $A$ of $Z(Common-Cold)$ diagnosed by $E1$}\label{ComA}
\end{figure}

The first component $A$ of $Z(Common-Cold)$, $A = (0.93,0.98,1.0,1.0;1.0)$ shown in Fig. \ref{ComA}. The procedure of the ranking score of $A$ is shown as follows:
$$
x_A = \frac{{\int_{0.93}^{0.98} {x\frac{{x - 0.93}}{{0.98 - 0.93}}dx + \int_{0.98}^1 {x\frac{{1 - x}}{{1 - 0.98}}dx} } }}{{\int_{0.93}^{0.98} {\frac{{x - 0.93}}{{0.98 - 0.93}}dx + \int_{0.98}^1 {\frac{{1 - x}}{{1 - 0.98}}dx} } }} = 0.97,
$$
$$h_A = 1,$$
$$STD_A = \sqrt {\frac{{\sum\nolimits_{j = 1}^4 {{{({x_j} - 0.9775)}^2}} }}{{4 - 1}}} = 0.033,$$
$$H(A)={w_1}{x_A} + {w_2}{h_A} + {w_3}\frac{1}{{1 + ST{D_A}}} = 0.9813,$$
namely, $H_{Z(Common-Cold)}(A) = 0.9813.$

In the same way, the ranking scores of all components can be calculated, shown as follows:
%$Z(Common-Cold) = (A,B)\\= ((0.93,0.98,1.0,1.0;1.0),(0.93,0.98,1.0,1.0;1.0)),$
%
%$Z(Meningitis) = (A,B) \\= ((0.04,0.1,0.18,0.23;1.0),(0.93,0.98,1.0,1.0;1.0)),$
%
%$Z(Measles) = (A,B) \\= ((0,0,0,0;1.0),(0.93,0.98,1.0,1.0;1.0)).$

$$H_{Z(Common-Cold)}(B) = 0.9813,$$
$$H_{Z(Meningitis)}(A) = 0.5101,$$
$$H_{Z(Meningitis)}(B)=0.9813,$$
$$H_{Z(Measles)}(A) = 0.4460,$$
$$H_{Z(Measles)}(B) =0.9813.$$

\item[Step 2:] Calculate respectively the deviation degree of diagnosis based on Eq. (\ref{D}). The deviation degree of the diagnoses of $E1$ can be obtained as follows:
$$
\begin{array}{ll}
 D(Common - cold) &=  \sqrt {\frac{{{0.7}\times{{(0.9813 - 1)}^2} + {0.3}\times{{(0.9813 - 1)}^2}}}{{{0.7}\times{{(0.446 - 1)}^2} + {0.3}\times{{(0.446 - 1)}^2}}}}\\
      &= 0.0338,
\end{array}
$$
$$D(Meningitis) = 0.7401,$$
$$D(Measles) = 0.8368.$$

%Based on the Equation. (\ref{S}), the similarity measure can be calculated as follows:
%$$S(Common - cold) = 1-0.0338 = 0.9662,$$
%$$S(Meningitis) = 1-0.7401 = 0.2599,$$
%$$S(Measles) = 1-0.8368 = 0.1632.$$

%This paper define that
%$$
%\begin{array}{l}
%S(Common - cold,Meningitis,Measles)\\
% = 1 - max(S(Common - cold),S(Meningitis),S(Measles))\\
% = 1 - 0.9662 = 0.0338
%\end{array}$$
\item[Step 3:] Generate BPAs of diagnoses. Firstly, the similarity measure between diagnosis and the maximal reference number is calculated based on Eq. (\ref{S}). The similarity measure between the diagnosis of $E1$ and the maximal reference number can be shown as follows:
$$S(Common - cold) = 1-D(Common - cold)= 1-0.0337 = 0.9662,$$
$$S(Meningitis) = 1-D(Meningitis)= 1-0.7402 = 0.2599,$$
$$S(Measles) = 1-D(Measles)= 1-0.8369 = 0.1632.$$
This paper define that
$$
\begin{array}{l}
S(Common - cold,Meningitis,Measles)\\
 = 1 - max(S(Common - cold),S(Meningitis),S(Measles))\\
 = 1 - 0.9662 = 0.0338.
\end{array}
$$

Then, the obtained similarity measure $S$ are normalized to obtain the BPAs of the diagnoses of $E1$, shown as follows:
$$m(Common - cold) = \frac{{0.9662}}{{0.9662 + 0.2599 + 0.1632 + 0.0338}} = 0.6789,$$
$$m(Meningitis) = \frac{{0.2599}}{{0.9662 + 0.2599 + 0.1632 + 0.0338}} = 0.1826,$$
$$m(Measles) = \frac{{0.1632}}{{0.9662 + 0.2599 + 0.1632 + 0.0338}} = 0.1147,$$
$$m(Common - cold,Meningitis,Measles) = 0.0238.$$
%\begin{equation*}
%\begin{array}{ll}
%m(Common - cold,Meningitis,Measles)&\\
% = \frac{{0.0337}}{{0.9663 + 0.2598 + 0.1631 + 0.0337}}&\\
% = 0.0237&,
%\end{array}
%\end{equation*}
%$m(Common - cold,Meningitis,Measles) = \frac{{0.0337}}{{0.9663 + 0.2598 + 0.1631 + 0.0337}} = 0.0237,$

In the same way, the BPAs of diagnoses of $E2$ and $E3$ can be calculated, as shown in Table \ref{TBt}.

\item[Step 4:] Based on Eq. (\ref{DS}), the BPAs obtained in Step $3$ are combined by applying Dempster's combination rule. The fusing results are shown in Table \ref{Tfr2}.
   % $$m(Common-cold)=0.7085,$$
%    $$m(Meningitis)=0.1076,$$
%    $$m(Measles)=0.1814,$$
%    $$m(Common-cold, Meningitis, Measles)=0.0025.$$

%\begin{table}
%\centering
% \caption{The BPAs of diagnoses from $E1$} \label{TBD}
% \tiny
% \begin{tabular}{ccccc}
%   \hline
%   \toprule
%   % after \\: \hline or \cline{col1-col2} \cline{col3-col4} ...
%    & m(Common-cold) & m(Meningitis) & m(Measles) & m(Common-cold, Meningitis, Measles) \\
%    \midrule
%   $E1$ & 0.6789 & 0.1826 & 0.1147 & 0.0238 \\
%   \bottomrule
%   \hline
% \end{tabular}
% \end{table}

\begin{table}
\centering
 \caption{The BPAs of diagnoses from three experts} \label{TBt}
 \tiny
 \begin{tabular}{ccccc}
   \hline
   \toprule
    & m(Common-cold) & m(Meningitis) & m(Measles) & m(Common-cold, Meningitis, Measles) \\
    \midrule
   $E1$ & 0.6789 & 0.1836 & 0.1147 & 0.0238 \\
   $E2$ & 0.4746 & 0.1674 & 0.1718 & 0.1862 \\
   $E3$ & 0.1717 & 0.1675 & 0.5596 & 0.1012 \\
   \bottomrule
   \hline
 \end{tabular}
 \end{table}

\begin{table}
\centering
 \caption{The fusing results by Dempster's rule of combination} \label{Tfr2}
 \tiny
 \begin{tabular}{ccccc}
   \hline
   \toprule
    & m(Common-cold) & m(Meningitis) & m(Measles) & m(Common-cold,Meningitis,Measles) \\
    \midrule
    & & & &\\
   fusing result & 0.7085 & 0.1076 & 0.1814 & 0.0025 \\
   \bottomrule
   \hline
 \end{tabular}
\end{table}

%\begin{figure}[ht!]
%  \centering
%  % Requires \usepackage{graphicx}
%  \includegraphics[scale=1]{fusing2.eps}\\
%  \caption{The visualization of fusing results}\label{Tfus2}
%\end{figure}

\end{description}
From the linguistic values of diagnoses in Table \ref{LV}, it can be known that both $E1$ and $E2$ tend to consider that the clinical patient is most likely to suffer from the common cold, and is less likely to suffer from measles. However, $E3$ tends to consider that the clinical patient is most likely to suffer from the measles, and is less likely to suffer from the common cold, which conflicts with the diagnoses of $E1$ and $E2$. In Table \ref{TBt}, it can be seen that the obtained BPAs are consistent with the above analysis.

The proposed method addresses the multiple sources and conflicting information by using Dempster's combination rule, and the fusing results are shown in Table \ref{Tfr2}. From Table \ref{Tfr2}, it can be seen that the final result of diagnosis is that the patient had a common cold, which is consistent with the truth that two of three experts consider that the patient is suffer from the common cold. From the obtained result of diagnosis, it can be seen that this method can solve the issues of the uncertainty and the confliction of information and can achieve a reasonable medical diagnosis.

To illustrate the effectiveness of Z-number modeling in medical diagnosis, in the following, we will give a case that the reliability of information is not considered. Now, all second components of Z-numbers are highest value $1$, that is $B=(1,1,1,1;1)$, and the first components are the same as Table \ref{Dot}.

The new diagnoses are presented in Table \ref{Newd}. The new BPAs of diagnoses from three experts are shown in Table \ref{TNBt}, and the new fusing results by Dempster's rule of combination are given in Table \ref{Tnfr}.

\begin{table}[ht!]
 \centering
 \caption{The new diagnoses of three experts are represented by linguistic terms} \label{Newd}
    \tiny
 \begin{tabular}{ll}
   \hline
   \toprule
   expert & Z-numbers represented by linguistic terms \\
   \midrule
          &${Z_{11}} = ((0.93,0.98,1.0,1.0;1.0),(1.0,1.0,1.0,1.0;1.0))$ \\
   $E1$ &${Z_{12}} = ((0.04,0.1,0.18,0.23;1.0),(1.0,1.0,1.0,1.0;1.0))$ \\
    &${Z_{13}} = ((0,0,0,0;1.0),(1.0,1.0,1.0,1.0;1.0))$ \\
    \midrule
    & ${Z_{21}} = ((0.58,0.63,0.80,0.86;1.0),(1.0,1.0,1.0,1.0;1.0))$ \\
   $E2$ &${Z_{22}} = ((0.04,0.1,0.18,0.23;1.0),(1.0,1.0,1.0,1.0;1.0))$\\
    &${Z_{23}} = ((0.04,0.1,0.18,0.23;1.0),(1.0,1.0,1.0,1.0;1.0))$ \\
    \midrule
    &${Z_{31}} = ((0.04,0.1,0.18,0.23;1.0),(1.0,1.0,1.0,1.0;1.0))$ \\
   $E3$ &${Z_{32}} = ((0.04,0.1,0.18,0.23;1.0),(1.0,1.0,1.0,1.0;1.0))$ \\
    &${Z_{33}} = ((0.72,0.78,0.92,0.97;1.0),(1.0,1.0,1.0,1.0;1.0))$ \\
    \bottomrule
   \hline
 \end{tabular}
 \end{table}

\begin{table}
\centering
 \caption{The new BPAs of diagnoses from three experts} \label{TNBt}
 \tiny
 \begin{tabular}{ccccc}
   \hline
   \toprule
    & m(Common-cold) & m(Meningitis) & m(Measles) & m(Common-cold, Meningitis, Measles) \\
    \midrule
   $E1$ & 0.4868 & 0.3689 & 0.1302 & 0.0141 \\
   $E2$ & 0.1711 & 0.1711 & 0.1711 & 0.4867 \\
   $E3$ & 0.1147 & 0.1827 & 0.5957 & 0.1069 \\
   \bottomrule
   \hline
 \end{tabular}
 \end{table}

\begin{table}
\centering
 \caption{The new fusing results by Dempster's rule of combination} \label{Tnfr}
 \tiny
 \begin{tabular}{ccccc}
   \hline
   \toprule
    & m(Common-cold) & m(Meningitis) & m(Measles) & m(Common-cold,Meningitis,Measles) \\
    \midrule
    & & & &\\
   fusing result & 0.3423 & 0.3420 & 0.3122 & 0.0035 \\
   \bottomrule
   \hline
 \end{tabular}
\end{table}

%\begin{figure}[ht!]
%  \centering
%  % Requires \usepackage{graphicx}
%  \includegraphics[scale=1]{fusing3.eps}\\
%  \caption{The visualization of new fusing results}\label{Tvnf}
%\end{figure}

From Table \ref{Tnfr}, it is obvious that the results are different considering the reliability of information and without considering the reliability of information. When the reliability of information is considered, the fusing diagnosis suggests that the patient is most likely to catch a common-cold and is less likely to suffer from meningitis. When the reliability of information is not considered, the fusing diagnosis suggests that the patient is most likely to catch a common-cold and is less likely to suffer from measles. In fact, due to the characteristics of medical diagnosis, more fuzzy concept and more uncertainties, the reliability of information is very important. If only to rely on the function of the first component, that is, without considering second components, there may exist the limitation to appropriately describing reliability of information, which leads to the incorrect result. Therefore, using Z-number to model is more reasonable in some decision making environment such as medical diagnosis.

\section{Conclusions}
\label{Conclu}
Human health has received so serious threat that the medical diagnosis becomes a very worthy of study. Various methods have been introduced to solve the medical diagnosis problems. Due to the own characteristics of medicine, more fuzzy concept and more uncertainties, fuzzy numbers, which have the ability to deal with the fuzzy and uncertain information, can provide a wonderful method for studying such issues. Since the reliability of the information in the evaluation should be included, this paper proposed a useful medical diagnosis method based on Z-numbers, where both the restriction and the reliability of the information are taken into consideration.

%In this paper, firstly, Z-numbers are utilized to represent the relevant information such as diagnoses obtained from multiple sources of evidence, that is, decision-makers, and the maximal reference number of Z-numbers is taken. Secondly, the deviation degree of the test numbers such as diagnoses are gotten. Then, the BPAs of the test numbers are generated based on the obtained deviation degree. Finally, in order to solve the problem of incomplete information, the obtained BPAs are fused by Dempster's combination rule. Through the proposed method, the problems of the uncertainty and the confliction of information can be solved. From the result of the medical diagnosis, it can be seen that the proposed method provides a reasonable and effective method for medical diagnosis. This method can also be applied to multi-attribute decision making.

The advantages of the proposed method include:

1) Firstly, in fact, due to the own nature of decision-making problems, more fuzzy concept and more uncertainties, fuzzy number has been widely applied in decision-making, since it could reasonably model and describe the uncertain and fuzzy information. However, in the actual research, we found that reliability of information in some decision-making environment, is very important too. If only to rely on the function of the fuzzy numbers, there may exist the limitation to appropriately describing reliability of information. In order to solve this problem, Z-numbers are utilized to model and describe the diagnoses of decision-makers, and a novel ranking method for fuzzy numbers is proposed to get BPAs of the diagnoses.

2) Secondly, in the procedure of addressing Z-numbers, there is a problem, that is how to deal with the restriction and the reliability of Z-number. In order to address the problem, in the procedure of dealing with Z-numbers, we treat the Z-number as a whole, avoiding converting the second component to a crisp number, which can avoid the loss of information.

3) Thirdly, in the application of Z-number, ranking Z-number is an important issue. The proposed decision making methodology can effectively address the issue, since the method can obtain reasonable and effective results by comparing with the existing methods.

4) Fourthly, in the real decision making problems, the proposed method can obtain reasonable and effective results, since the proposed methodology uses Z-number as a whole to model, and generates BPAs by taking into account the different importance of two components of Z-number.

5) Finally, Dempster's combination rule is applied to obtain the final fusion results, which could make the medical diagnosis decision more adequate and reasonable. Compared with the existing methods, the proposed method provides a simple and effective method for medical diagnosis. At the same time, this method can also be applied to other multi-attribute decision making.

Although the proposed method is demonstrated to be effective through illustrative examples and in-depth discussion, there are still some aspects that can be improved in the future studies. At first, within the proposed method the difference in the importance between two components, namely assessment value $A$ and reliability measure $B$, of a Z-number is needed to be studied further. Intuitively, these two components should have different importance. But how to determine their weights is still an unsolved issue. In the paper, the OWA operator is employed to generate the weighing vector without any requirements for other information. Although this approach is not bad, there is still a practical demand to obtain the weights of components $A$ and $B$ in more reasonable and data-based ways. Second, in the proposed ranking approach of two Z-numbers, in order to get the similarity measure between test Z-number and reference Z-number, at present only positive ideal reference is taken into consideration. However, as shown in many typical multi-criteria decision making methods such as TOPSIS, the negative ideal reference may be also very important. Therefore, in the future research, the positive and negative ideal references will be investigated simultaneously to derive more reasonable results and better performance.

\begin{acknowledgements}
The authors greatly appreciate the reviewers' suggestions and the editors' encouragement. The work is partially supported by National Natural Science Foundation of China (Grant No. 61671384), Natural Science Basic Research Plan in Shaanxi Province of China (Program No. 2016JM6018), Aviation Science Foundation (Program No. 20165553036), the Fund of SAST (Program No. SAST2016083), the Seed Foundation of Innovation and Creation for Graduate Students in Northwestern Polytechnical University (Program No. Z2016122).
\end{acknowledgements}

%\begin{acknowledgements}
%\end{acknowledgements}

\bibliographystyle{spmpsci}      % mathematics and physical sciences
\bibliography{myreference}   % name your BibTeX data base

%\begin{thebibliography}{}
% etc
%\end{thebibliography}

\end{document}